\documentclass{article}
\usepackage{spconf,amsmath,graphicx}
\usepackage{multirow}
\usepackage{color}

\usepackage {setspace}
\usepackage{hyperref}
\hypersetup{hidelinks,
	colorlinks=true,
	allcolors=black,
	pdfstartview=Fit,
	breaklinks=true}
\title{Reference-guided texture and structure inference for image inpainting}
\name{Taorong Liu$^{1}$ \qquad Liang Liao$^{2}$ \qquad Zheng Wang$^{1}$ \qquad Shin'ichi Satoh$^{3}$}
\address{$^{1}$ School of Computer Science, Wuhan University, Wuhan, China \\$^{2}$ S-lab, Nanyang Technological University, Singapore\\      $^{3}$ National Institute of Informatics, Tokyo, Japan}
\begin{document}
%\ninept
%
\maketitle

\begin{abstract}
Existing learning-based image inpainting methods are still in challenge when facing complex semantic environments and diverse hole patterns. The prior information learned from the large scale training data is still insufficient for these situations. Reference images captured covering the same scenes share similar texture and structure priors with the corrupted images, which offers new prospects for the image inpainting tasks. Inspired by this, we first build a benchmark dataset containing 10K pairs of input and reference images for reference-guided inpainting. Then we adopt an encoder-decoder structure to separately infer the texture and structure features of the input image considering their pattern discrepancy of texture and structure during inpainting. A feature alignment module is further designed to refine these features of the input image with the guidance of a reference image. Both quantitative and qualitative evaluations demonstrate the superiority of our method over the state-of-the-art methods in terms of completing complex holes. Code is available at \href{https://github.com/Cameltr/RGTSI}{https://github.com/Cameltr/RGTSI}.
\end{abstract}


\section{Introduction}
\label{sec:intro}

\begin{figure}[tb]
	\centering
\footnotesize{
    \resizebox{\linewidth}{!}{
		\begin{tabular}{ccc}
			\includegraphics[width=0.31\columnwidth]{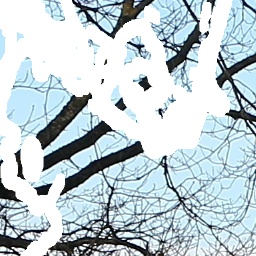} &
			\includegraphics[width=0.31\columnwidth]{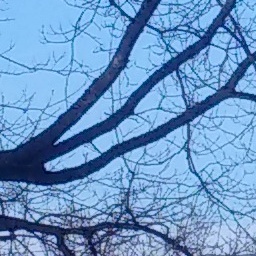} &
			\includegraphics[width=0.31\columnwidth]{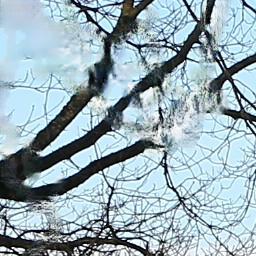} \\	
			(a) Input image & (b) Reference image & (c) MEDFE~\cite{MEDFE}\\	\includegraphics[width=0.31\columnwidth]{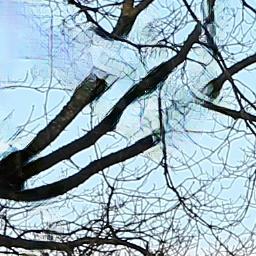} &			\includegraphics[width=0.31\columnwidth]{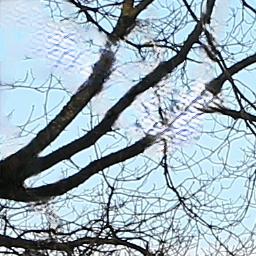} &			\includegraphics[width=0.31\columnwidth]{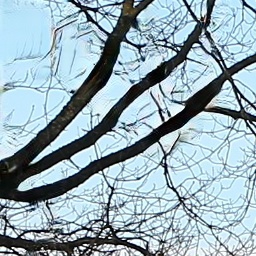}  \\
			(d) EC~\cite{EC} & (e) GI~\cite{GI} & (f) Ours\\
	\end{tabular}}}
	\vspace{-3mm}
   \caption{Visual comparison on the DPED10K dataset. Our method is more effective in reducing bending and blurring of tree branches.}
     \vspace{-3mm}

\label{fig:deepfeature}
\end{figure}

Image inpainting aims to fill up the missing parts of an image with plausible content, which can be used to repair corrupted areas or remove unwanted targets in photos, \emph{etc}. Its main idea is to find sufficient prior information, such as smoothness \cite{3} and self-similarity \cite{2,18}, as a reference to fill the holes. Recently, learning-based inpainting methods have attracted great attention. Pathak \emph{et al.} first introduced Context Encoders \cite{25}, followed which neural patch synthesis \cite{patch}, residual learning \cite{residual}, partial convolution \cite{PConv}, and others \cite{4,MEDFE,LiaoXWLS21} were proposed. To assist the complex image inpainting, structural priors were introduced to guide the inpainting task, including edges \cite{EC,PRVS,Liang}, contours \cite{Contour}, and segmentation maps \cite{27,Liao2021CVPR}. These methods show that the structural prior effectively helps to improve the quality of the completed images. In spite of this, they still struggle in cases when holes are large or the expected content contains complicated semantic information (\emph{e.g.}, Fig.~\ref{fig:deepfeature}(c), (d) and (e)).

\tabcolsep=0.5pt

To seek more prior information to help solve this type of ill-posed problems, reference images with similar textures and structures are introduced in some vision tasks, such as image super-resolution ~\cite{MASA,SRNNT}, image compression ~\cite{ho2021rr} and action recognition~\cite{zhongicassp}. Their motivation is to utilize rich textures from the references to compensate for the lost details in the input images and thereby produce more detailed and realistic content. Inspired by this, reference-guided image inpainting is introduced in this paper, which focuses on filling in missing areas guided by external references containing complete and similar structures and details. However, reference images covering the same scene are often captured under different conditions, resulting in different styles and geometries of the input and reference images (\emph{e.g.}, Fig.~\ref{fig:deepfeature}(a) and (b)).

In this paper, we propose a novel reference-based encoder-decoder network for image inpainting. Considering the pattern discrepancy of texture and structure for completion, such as large stylistic differences in texture and large geometric differences in structure, we introduce reference images as a guide to complete texture and structure features separately. Specifically, the input images and the reference images are first encoded to generate the respective texture features and structure features. Then, both features extracted from the input and reference images are aligned by a feature alignment module, and then fused in a multi-scale filling manner. A equalization operation as in \cite{MEDFE} is employed to ensure that the attentions in every channel are the same and the feature representations are consistent as well. The equalized features are concatenated to the decoder with the features from the encoder to generate the inpainting result. To promote this task, we construct a dataset named DPED10K, in which the reference images  generated based on the SIFT~\cite{SIFT} share similarities with the structure and texture of the input images.

The main contributions of this paper are: 1) We propose the idea of reference-guided inpainting to compensate for the insufficient priors in completing complex scenes. 2) We design a reference-based encoder-decoder network and a novel feature alignment module to finely align and fuse the feature of the input image and reference image. 3) We build a reference images dataset, to facilitate further research and performance evaluation of reference-guided image inpainting.

\begin{figure}[tb]
\centering
\begin{minipage}[b]{0.1\linewidth}
  \centering
  \centerline{\includegraphics[width=2cm]{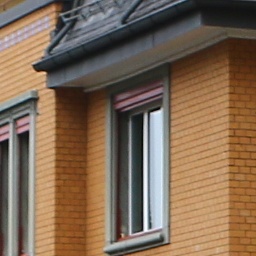}}
%  \vspace{2.0cm}
  \medskip
\end{minipage}
\hspace{10mm}
\begin{minipage}[b]{0.1\linewidth}
  \centering
  \centerline{\includegraphics[width=2cm]{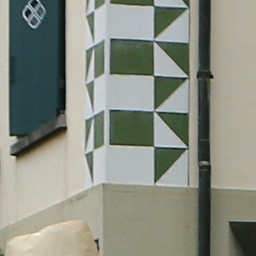}}
%  \vspace{2.0cm}
  \medskip
\end{minipage}
\hspace{10mm}
\begin{minipage}[b]{0.1\linewidth}
  \centering
  \centerline{\includegraphics[width=2cm]{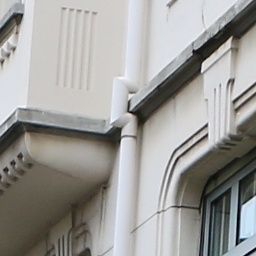}}
%  \vspace{1.5cm}
  \medskip
\end{minipage}
\hspace{10mm}
\begin{minipage}[b]{0.1\linewidth}
  \centering
  \centerline{\includegraphics[width=2cm]{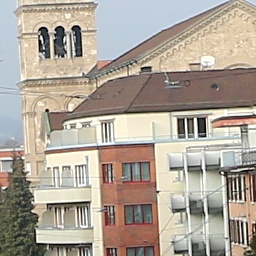}}
%  \vspace{1.5cm}
  \medskip
\end{minipage}

\begin{minipage}[b]{0.1\linewidth}
  \centering
  \centerline{\includegraphics[width=2cm]{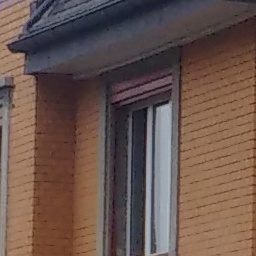}}
%  \vspace{1.5cm}
  \medskip
\end{minipage}
\hspace{10mm}
\begin{minipage}[b]{0.1\linewidth}
  \centering
  \centerline{\includegraphics[width=2cm]{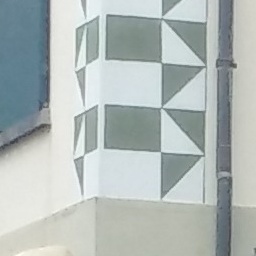}}
%  \vspace{1.5cm}
  \medskip
\end{minipage}
\hspace{10mm}
\begin{minipage}[b]{0.1\linewidth}
  \centering
  \centerline{\includegraphics[width=2cm]{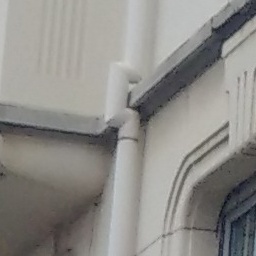}}
%  \vspace{1.5cm}
  \medskip
\end{minipage}
\hspace{10mm}
\begin{minipage}[b]{0.1\linewidth}
  \centering
  \centerline{\includegraphics[width=2cm]{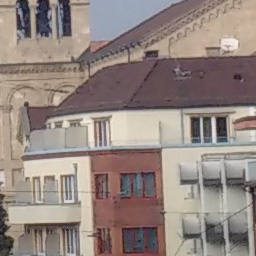}}
  \medskip
\end{minipage}
  \vspace{-0.5cm}

\caption{Examples from the DPED10K dataset. The first row is the input image and the second row is its corresponding reference image. Each pair of the input image and reference image shares the similarities of structure and texture.}
\label{fig:dataset}
\vspace{-3mm}
\end{figure}

\section{Approach}

% As illustrated in Fig \ref{fig:pipeline}, We use a reference-based encoder-decoder to jointly learn the structure feature and texture features of the image and its reference image, and align them for feature fusion. In this section, we detailedly describe the RBED, the feature alignment module, and the loss functions.

\begin{figure*}[t]
    \centering
    \includegraphics[width=15cm]{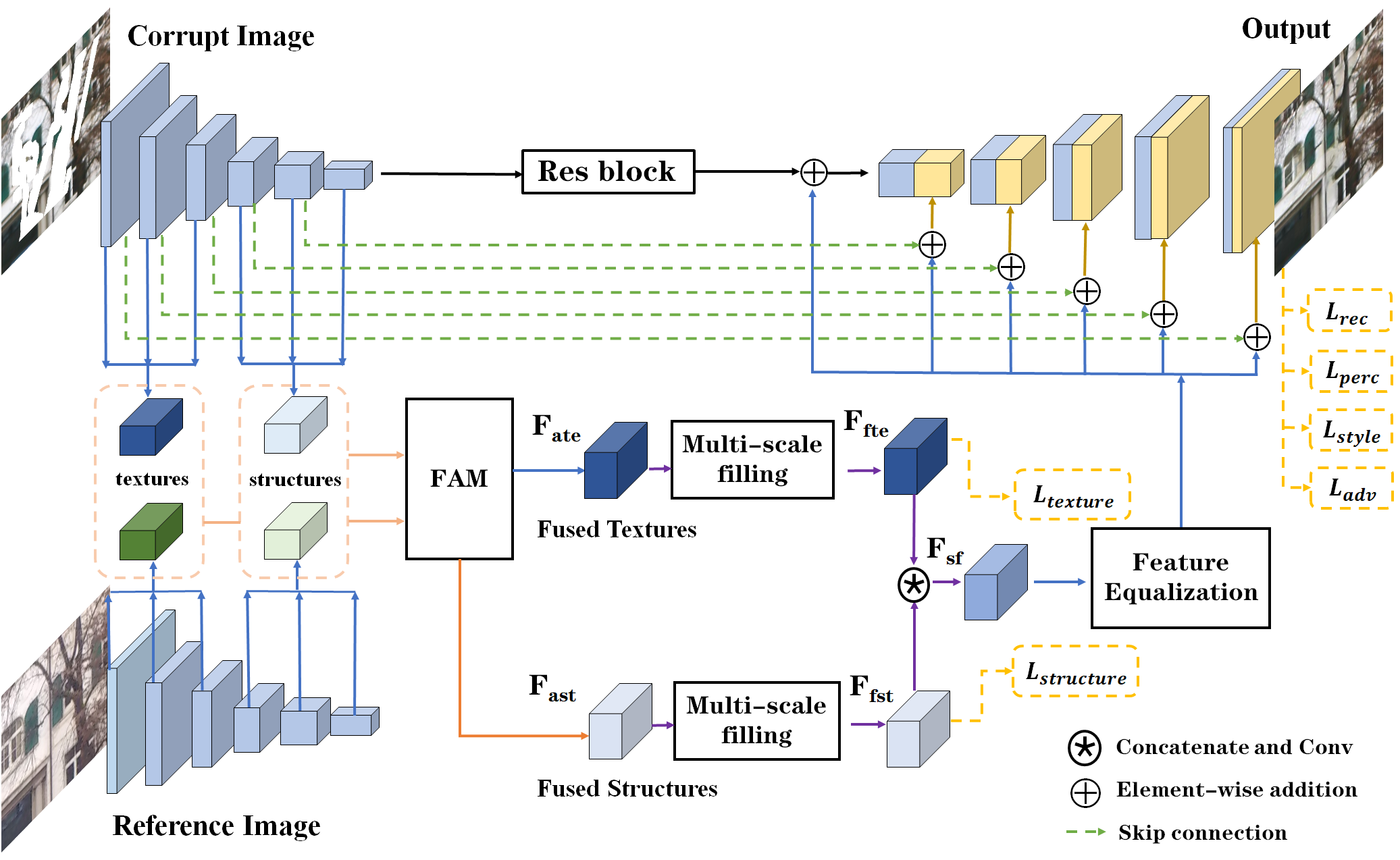}
    \vspace{-2mm}
    \caption{The overview of the proposed pipeline. We adopt a referenced-based encoder-decoder to jointly fill image holes. The features of the input and reference images are aligned and fused by the Feature Alignment Module (FAM). We recover the holes in multi-scale within the aligned features and equalize the output features. The equalized features contain consistent structure and texture features and are supplemented to the decoder by skip connections to generate the output image.
}
    \label{fig:pipeline}
        \vspace{-3mm}

\end{figure*}
\label{sec:method}

\subsection{Dataset}
\label{sec:dataset}
For image inpainting problems based on the reference image, the similarity between the image and its reference image are of great significance to inpainting results. However, to the best of our knowledge, we have not found such a publicly available dataset. Therefore, we construct such a dataset based on DPED ~\cite{DPED} dataset that consists of real photos captured by three different phones and one high-end reflex camera. We select images captured via Blackberry and Sony and generate an input-reference dataset based on SIFT~\cite{SIFT} feature matching, which characterizes local texture and structure pattern that is in line with the objective of local texture and structure matching.

For each image pair, we randomly crop 256$\times$256 patches from images as input images, while the reference images are cropped from the corresponding images. In this way, we collect 10K paired patches as a training dataset and 200 pairs are randomly collected as a test dataset. Some image samples are shown in Fig. \ref{fig:dataset}. We refer to the collected dataset as DPED10K, which has be released to facilitate the research on reference-guided image inpainting.

\subsection{Referenced-Based Encoder-Decoder}
The network architecture is illustrated in Fig.~\ref{fig:pipeline}, which consists of two encoders that share the same parameters and one decoder to fill the missing regions in an end-to-end training manner. Meanwhile, we set the residual blocks between the encoder and decoder with dilated convolutions, which can well increase the receptive field size to cover more contextual information.

%Specifically, for the encoder side of the input image, we denote its structure features as $F_{ist}$ and its texture features as $F_{ist}$. Similarly, for the reference image, we denote its structure features as $F_{rst}$ and its texture features as $F_{rte}$. 

In the encoder, we refer to MEDFE~\cite{MEDFE} to reorganize the shallow features of the encoder into texture features and the deep features into structure features. Then the texture and structure features from different convolutional layers are resized to the same size and concatenated, respectively. To refine the features of input corrupted images by that of reference images, we propose a Feature Alignment Module (FAM) based on deformable convolution~\cite{Deform} and dynamic offset estimator~\cite{SSEN}. The process of FAM is formulated as follows: 
\begin{equation}
\begin{aligned}
    F_{ate} = G(F_{ite}, F_{rte}),\\
    F_{ast} = G(F_{ist}, F_{rst}).
\end{aligned}
\end{equation}
where $G(\cdot)$ denotes the feature alignment module. $F_{ite}$, $F_{rte}$, $F_{ist}$ and $F_{rst}$ are the texture features and the structure features from the input and reference images. $F_{ate}$ and $F_{ast}$ are the aligned and fused structure and texture features, respectively. FAM aims to fuse the guidance information from the reference image to help filling missing area in input image.

After the features are aligned and fused, we design the structure branch and the texture branch accordingly to fill holes for $F_{ate}$ and $F_{ast}$. Each branch with several streams consists of different kernel sizes of partial convolutions ~\cite{PConv} to fill holes in multiple scales. To ensure the hole filling focuses on the textures and structures, we utilize the ground-truth image and its structure image generated by RTV~\cite{RTV} to supervise the output of two branches $F_{fst}$ and $F_{fte}$. After that, we concatenate $F_{fte}$ and $F_{fst}$ and then generate $F_{sf}$ by simple fusion of $1\times1$ convolutional layers. The texture and structure representations in $F_{sf}$ after feature equalization are supplemented to different feature layers with skip connections to generate the output image.
%We use $1 \times 1$ convolution to separately map $F_{fst}$ and $F_{fte}$ to the color image $I_{ost}$ and the color image $I_{ote}$, respectively. 
%The pixel-wise $L1$ loss can be written as follows:
%\begin{equation}
%\begin{aligned}
    %L_{recst} = \parallel I_{ost} - I_{st}\parallel_1,\\
    %L_{recte} = \parallel I_{ote} - I_{gt}\parallel_1,
%\end{aligned}
%\end{equation}
%$I_{gt}$ is the ground truth image and $I_{st}$ is the structure image of $I_{gt}$. We use an edge-preserving smoothing method RTV \cite{RTV} to generate $I_{st}$. 

\subsection{Feature Alignment Module}
\begin{figure}[t]
    \centering
    \includegraphics[width=7.5cm]{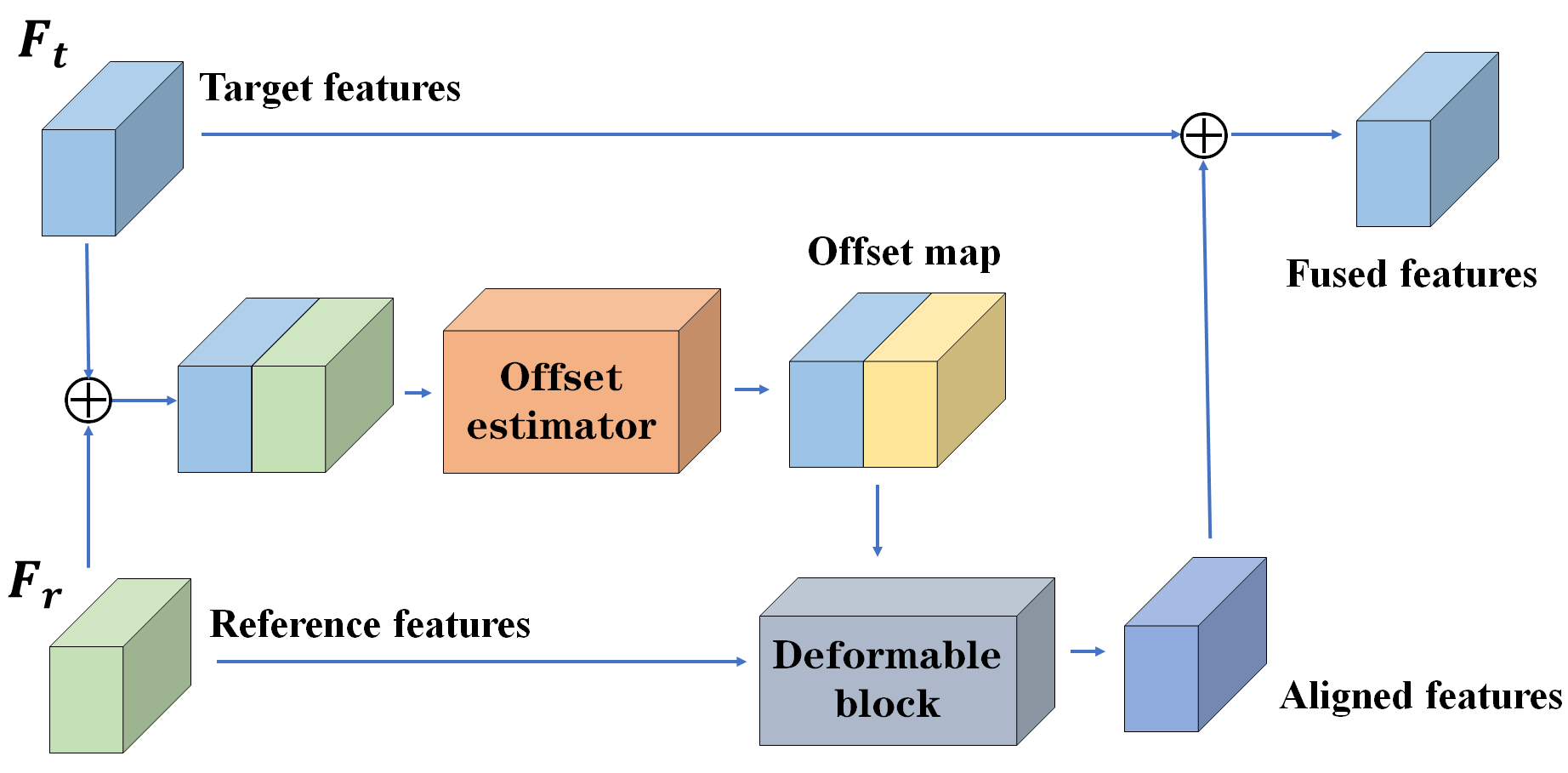}
    \vspace{-0.3cm}
    \caption{Illustration of the Feature Alignment Module.}
    \label{fig: FAM}
    \vspace{-0.4cm}
\end{figure}

In this paper, we use the deformable convolution~\cite{Deform}, which models deformation by learning an additional offset, to achieve feature-level alignment. This offset enables the model to focus on the target region when extracting features by shifting its convolution kernels. Formally, the deformable convolution operation is defined as follows: 
\begin{equation}
    y(p_0) = \sum_{n=1}^{N} \omega_n \cdot X(p+p_n+\Delta p_n),
\end{equation}
where $X$ is the input, $Y$ is the output, and $n$ and $N$ denote the number of indexes and the number of kernel weights, respectively. $\omega_n$, $p$, $p_n$, and $\Delta p_n$ are the $n$-th kernel weight, the central index, the $n$-th fixed offset and the learnable offset for the $n$-th position, respectively. 

As shown in Fig.~\ref{fig: FAM}, FAM first connects the input target feature $F_t$ with the reference feature $F_r$. Then, the embedded features are passed to the dynamic offset estimator~\cite{SSEN} to generate the offset map between the input feature and reference feature. The dynamic offset estimator can capture the similarity located from near to far distances. Guided by the offset map, the reference features are aligned by deformable convolution~\cite{Deform}. It is important to note that the offset estimator~\cite{SSEN} can be utilized for single image inpainting if there is no reference image.

\tabcolsep=0.7pt
\begin{figure*}[htb]
	\centering
\footnotesize{
		\begin{tabular}{cccccccc}
			
			\includegraphics[width=0.125\textwidth]{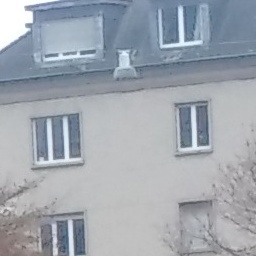} &
			\includegraphics[width=0.125\textwidth]{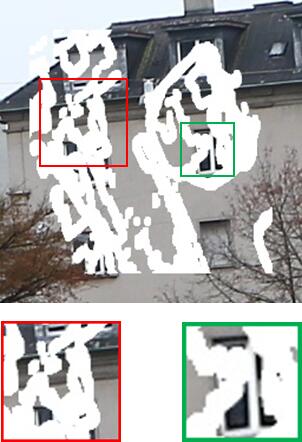} &	\includegraphics[width=0.125\textwidth]{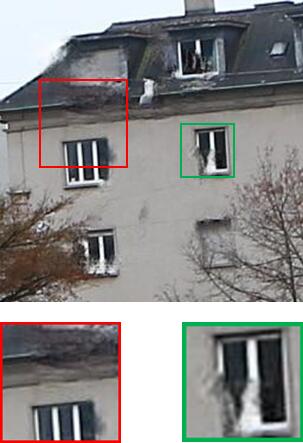} &	\includegraphics[width=0.125\textwidth]{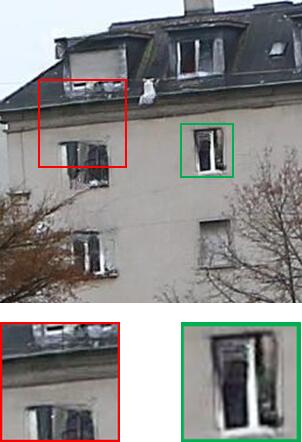} &  
			\includegraphics[width=0.125\textwidth]{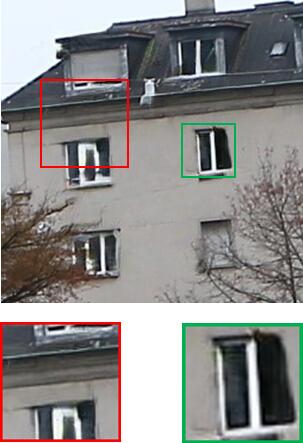} &
			\includegraphics[width=0.125\textwidth]{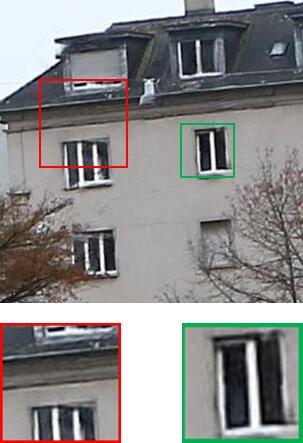} &
			\includegraphics[width=0.125\textwidth]{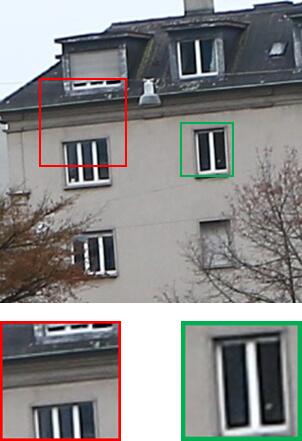}  &\\
			
			\includegraphics[width=0.125\textwidth]{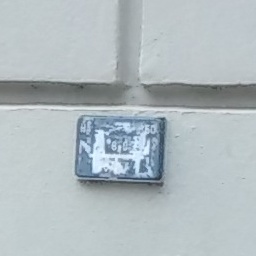} &
			\includegraphics[width=0.125\textwidth]{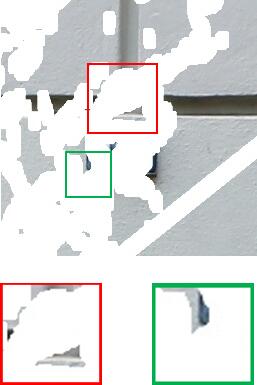} &	\includegraphics[width=0.125\textwidth]{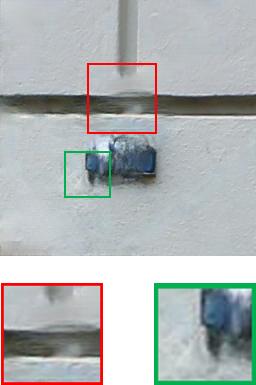} &	\includegraphics[width=0.125\textwidth]{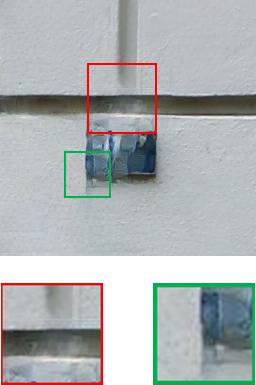} &  
			\includegraphics[width=0.125\textwidth]{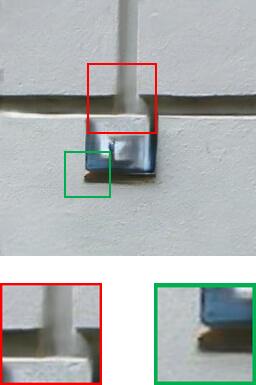} &
			\includegraphics[width=0.125\textwidth]{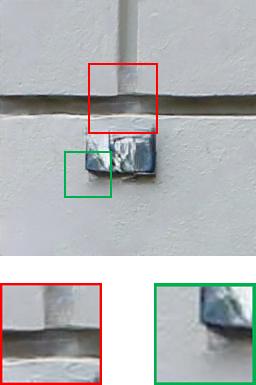} &
			\includegraphics[width=0.125\textwidth]{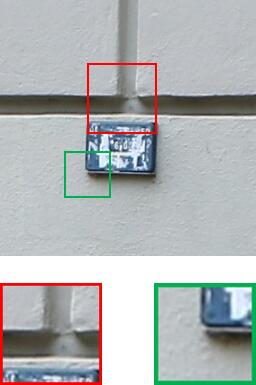}  &\\
			(a) Reference & (b) Input & (c) MEDFE~\cite{MEDFE} & (d) EC~\cite{EC} & (e) GI~\cite{GI} & (f) Ours & (g) Ground-truth & \\
	\end{tabular}}
	\vspace{-0.2cm}
   \caption{Qualitative comparisons of inpainting results on image samples from DPED10K dataset.}
     \vspace{-0.1cm}
\label{fig:qualitive}
\end{figure*}

\begin{table*}[tb]
\centering
\setlength\tabcolsep{11pt}
\resizebox{1.0\textwidth}{!}
{
\begin{tabular}{c|ccccc|c}
\hline
   \multirow{2}{*}{Method}   & \multicolumn{5}{c|}{Mask of Ratio}  &\multirow{2}{*}{Average}       \\\cline{2-6}
 ~  & 10\%-20\%     & 20\%-30\%         & 30\%-40\%         & 40\%-50\%         & 50\%-60\%             & ~ \\
\hline
 MEDFE~\cite{MEDFE}                      & 29.88/0.957   & 26.45/0.912       & 23.86/0.854       & 22.05/0.787       & 19.44/0.638           & 24.33/0.829               \\
 EC~\cite{EC}                        & 29.30/0.950   & 26.37/0.909       & 24.23/0/858       & 22.40/0.794       & 20.10/0.668           & 24.48/0.836               \\
 GI~\cite{GI}                         & 31.24/0.965   & 27.34/0.927       & 24.86/0.877       & 22.67/0.821       & 19.88/\textbf{0.694}  & 25.20/0.857               \\
 \hline
 No reference               & 31.35/0.966   & 27.87/0.933       & 25.36/0.885       & 23.20/0.826       & 20.22/0.688           & 25.60/0.860               \\
 Random reference           & 31.44/\textbf{0.967}   & 27.95/0.933       & 25.42/0.887       & 23.32/0.828       & 20.36/0.690           & 25.70/0.861               \\
 Ours (reference)            & \textbf{31.77/0.967}   & \textbf{28.07/0.935}     & \textbf{25.63/0.888}       & \textbf{23.49/0.832}     & \textbf{20.58/0.694}   & \textbf{25.91/0.863} \\
\hline
\end{tabular}
}
\vspace{-0.4cm}
\caption{PSNR/SSIM quantitative comparisons of ours against other baselines and ablation study of the impact of reference image on DPED10K dataset.}
    
\label{table1}
\vspace{-0.5cm}
\end{table*}

\section{Experiment}
\label{sec:experiment}
We evaluate our method on the proposed DPED10K. Irregular masks are obtained from \cite{PConv} and classified based on their hole sizes relative to the entire image. The resolution of all images and corresponding masks during training and testing is resized to $256\times256$. The model is implemented in PyTorch and optimized with the Adam optimizer with a learning rate of $2\times10^{-4}$ and a batch size of 1 on a single Tesla V100 GPU. The training process of our proposed model are stopped after 80 epochs. We fuse the reconstruction loss, perceptual loss~\cite{perceptual}, style loss, and Relativistic Average LS Adversarial loss~\cite{relativistic} to supervise the training process. We compare our method with three state-of-the-art methods: MEDFE~\cite{MEDFE}, EC \cite{EC}, GI \cite{GI}. %In this section, both quantitative and qualitative comparisons are conducted to demonstrate the advantages of the proposed RBED.


% \usepackage{multirow}

\textbf{Qualitative Comparison.} For a fair evaluation, we conduct experiments on filling irregular holes in the input images. The qualitative results are shown in Fig.~\ref{fig:qualitive}. Generally, these baselines are challenging to recover the image content effectively. For example, in the first row of Fig.~\ref{fig:qualitive}(c), (d), and (e), the window appears to have broken and blurred lines. In contrast, our method can recover the structure and texture well by introducing a reference image, and Fig.~\ref{fig:qualitive}(f) show that our method can effectively generate more pleasing content.

\textbf{Quantitative Comparison.} We conduct quantitative evaluations on the DPED10K dataset with different mask ratios. For the evaluation metrics, we use two common metrics: SSIM and PSNR. The comparison results of filling the irregular holes are shown in Table.\ref{table1}. Our method can achieve better performance in filling irregular holes at all hole image ratios. We also compared the complexity of our proposed model with these methods. The average execution time of model to complete an image of $256 \times 256$ is 77ms, compared to 40ms for MEDFE ~\cite{MEDFE}, 22ms for EC ~\cite{EC}, and 106ms for GI ~\cite{GI} respectively.


%\subsection{Ablation Study}

\textbf{Impact of reference image.} To prove the effectiveness of the reference images, we set up two comparison experiments(\emph{i.e.}, No reference and Random reference). Specifically, the experiment without reference uses a solid black $256 \times 256$ image, while the experiment with a random reference disrupts the order of the reference images. Results are shown in Table~\ref{table1}. Since FAM can be utilized for single image inpainting when there is no reference image, results without reference still outperform well than other baselines. For the experiment without reference, the corrupted input image lacks the information of structure and texture feature of the reference image, and can only be repaired by learning the texture and structure information around its holes. The experiment with a random reference can still provide a small amount of feature information for the inpainting result after the feature alignment process. Therefore, the performance can be better than the experiment without reference.

\section{Conclusion}
\label{sec:conclusion}
In this paper, we proposed an end-to-end network structure that utilizes a complete reference image with a similar texture and structure to the input image. To address the problem of fusing the feature of the input image and reference image, we proposed a feature alignment module that can finely align and fuse the feature at the pixel level finely. Both quantitative and qualitative experiments are conducted to demonstrate the effectiveness of our proposed methods. In addition, a new dataset DPED10K is constructed to facilitate the evaluation of reference-guided inpainting methods and provides a benchmark for future reference-guided inpainting research.

\vspace{3mm}
\noindent \textbf{Acknowledgement:}
This work was supported by National Natural Science Foundation of China under Grant 62171325.

\clearpage

% Below is an example of how to insert images. Delete the ``\vspace'' line,
% uncomment the preceding line ``\centerline...'' and replace ``imageX.ps''
% with a suitable PostScript file name.
% -------------------------------------------------------------------------

%\vfill
%\pagebreak


\end{document}